\documentclass[conference]{IEEEtran}
\IEEEoverridecommandlockouts
\usepackage{cite}
\usepackage{amsmath,amssymb,amsfonts}
\usepackage{algorithmic}
\usepackage{graphicx}
\usepackage{textcomp}
\usepackage{xcolor}
\usepackage{tabularx}
\usepackage{enumerate}
\usepackage{booktabs}
\usepackage{subfigure}

\newcolumntype{C}[1]{>{\arraybackslash}m{#1}}

\def\BibTeX{{\rm B\kern-.05em{\sc i\kern-.025em b}\kern-.08em
    T\kern-.1667em\lower.7ex\hbox{E}\kern-.125emX}}
\begin{document}

\title{Autonomous AI-enabled Industrial Sorting Pipeline for Advanced Textile Recycling \\
}

\makeatletter
\newcommand{\linebreakand}{%
  \end{@IEEEauthorhalign}
  \hfill\mbox{}\par
  \mbox{}\hfill\begin{@IEEEauthorhalign}
}
\makeatother

\author{\IEEEauthorblockN{1\textsuperscript{st} Yannis Spyridis}
\IEEEauthorblockA{\textit{Department of Networks and} \\ \textit{Digital Media,} \\
\textit{Kingston University,}\\
London, UK \\
yannis.spyridis@kingston.ac.uk}
\and
\IEEEauthorblockN{2\textsuperscript{nd} Vasileios Argyriou}
\IEEEauthorblockA{\textit{Department of Networks and} \\ \textit{Digital Media,} \\
\textit{Kingston University,}\\
London, UK \\
vasileios.argyriou@kingston.ac.uk}
\and
\IEEEauthorblockN{3\textsuperscript{rd} Antonios Sarigiannidis}
\IEEEauthorblockA{\textit{K3Y Ltd,} \\
Sofia, Bulgaria \\
asarigia@k3y.bg}
\linebreakand
\IEEEauthorblockN{4\textsuperscript{th} Panagiotis Radoglou}
\IEEEauthorblockA{\textit{K3Y Ltd,} \\
Sofia, Bulgaria \\
pradoglou@uowm.gr}
\and
\IEEEauthorblockN{5\textsuperscript{th} Panagiotis Sarigiannidis}
\IEEEauthorblockA{\textit{Department of Informatics and} \\ \textit{Telecommunications Engineering,} \\
\textit{University of Western Macedonia,}\\
Kozani, Greece \\
psarigiannidis@uowm.gr}
}

\maketitle

\begin{abstract}
    The escalating volumes of textile waste globally necessitate innovative waste management solutions to mitigate the environmental impact and promote sustainability in the fashion industry. This paper addresses the inefficiencies of traditional textile sorting methods by introducing an autonomous textile analysis pipeline. Utilising robotics, spectral imaging, and AI-driven classification, our system enhances the accuracy, efficiency, and scalability of textile sorting processes, contributing to a more sustainable and circular approach to waste management. The integration of a Digital Twin system further allows critical evaluation of technical and economic feasibility, providing valuable insights into the sorting system's accuracy and reliability. The proposed framework, inspired by Industry 4.0 principles, comprises five interconnected layers facilitating seamless data exchange and coordination within the system. Preliminary results highlight the potential of our holistic approach to mitigate environmental impact and foster a positive shift towards recycling in the textile industry.
\end{abstract}


\section{Introduction}

Textile waste has become a pressing global concern, with today's overconsumption and throwaway cultures contributing significantly to escalating volumes of discarded textiles. The global clothing and footwear consumption is expected to surpass 100 million tonnes by 2030, while in the European Union, approximately 5.8 million tonnes of textile are discarded every year \cite{eea2019}. The environmental impact of this waste is detrimental, including consequences in resource depletion and pollution, that underscore the urgent need for innovative waste management solutions.

Leading cause for this problem, is the linear production and consumption model that has been traditionally associated with the fashion industry \cite{mishra2021anatomy}. This approach has significantly contributed to the textile waste crisis, as it lacks integration of effective recycling processes. The problem has been further amplified by the rapid evolution of fashion trends, which inevitably leads to shorter garment lifespans. While attempts to limit this crisis have been made, current methods that rely on manual sorting are inefficient, labor-intensive, and often prone to errors, thus hindering the transition towards a more sustainable fashion industry \cite{mukendi2020sustainable}.

Besides the direct environmental concerns of textile waste, the magnitude of the problem also presents a missed opportunity for significant resource recovery and development of circular economy practices. To address this issue, a new paradigm of automated textile analysis pipelines is required, integrating automation, artificial intelligence (AI), and computer vision into the basis of textile recycling systems \cite{sikka2022artificial}. Therefore, in this work our aim is to tackle the inefficiencies in traditional textile sorting by introducing a state-of-the-art autonomous textile analysis pipeline. 

Our system utilises robotics, spectral imaging, and AI-driven classification, to enhance the accuracy, efficiency, and scalability of textile sorting processes, thus contributing to a more sustainable and circular approach to textile waste management. Notably, our approach is complemented by a Digital Twin (DT) system, that enhances the associated processes by allowing the critical evaluation of their technical and economic feasibility. Moreover, through the integration of data from the AI-driven models, the DT provides valuable insights into the accuracy and reliability of the sorting system. Through this holistic approach, we aim to mitigate the environmental impact of fabric waste and assist in the industry's shift towards a more positive approach to recycling.

In addition, our autonomous textile analysis pipeline incorporates laser segmentation to optimise the production of textile fractions tailored for diverse recycling processes. This approach allows precise cutting of materials, enabling the removal of components such as buttons and zippers, thus enhancing the efficiency of the sorting system by ensuring that the resultant fractions are well-suited for repurposing. This targeted segmentation further contributes to the development of a more sustainable and resource-efficient textile waste management system. 

Comprising five interconnected layers, the architectural framework of our introduced pipeline allows the seamless exchange of data and coordination within the system. This interconnected approach in conjunction with the Digital Twin  insights, leads to the development of the desired sustainability in our model, aligned with the principles of Industry 4.0. 

The remainder of this paper is structured as follows: Section \ref{sec:background} explores related work on textile analysis. Section \ref{sec:methodology} presents the methodology of our approach by describing the system architecture, subsystems, and the AI models that drive the automated textile analysis pipeline. Section \ref{sec:evaluation} presents the results of the AI-driven process and highlight the role of the Digital Twin towards the feasibility analysis. Finally, section \ref{sec:conclusion} summarises key findings and emphasises the impact of our system.

\section{Background}
\label{sec:background}
\subsection{Textile Analysis}

Textile analysis systems encompass a wide range of techniques and methods for evaluating various aspects of textiles. These systems typically involve both objective and subjective testing methods, including the use of instrumental techniques and multi-analytical approaches to study and identify textile component materials, such as spectroscopy \cite{zhou2019textile} and chemometrics \cite{peets2017identification}. To address the registration challenge arising from the utilisation of multiple sensors with diverse modalities in spectral analysis, subpixel solutions, due to resolution variations have been introduced \cite{argyriou2005performance, argyriou2010sub, argyriou2003sub}. Other approaches to detect and recognise textile materials are based on photometric techniques \cite{argyriou2010photometric} offering precise visualisation of the fabric patterns. Textile analysis plays a vital role in evaluating fabric waste generation and conservation methods, including the qualitative and quantitative analysis of post-consumer textile waste \cite{stanescu2021state}.

State-of-the-art textile classification methods involve the use of spectral analysis and AI for accurate categorisation of textile samples. Spectroscopic techniques in particular, can rely on near-infrared spectroscopy and pattern recognition methods such as soft independent modelling of class analogy, least squares support machines, and extreme learning machines, for classifying textile fabrics \cite{sun2016classification}. In addition, the associated spectra can be analysed through statistical multivariate methods, achieving high speed and accuracy in classification \cite{riba2020circular}, thus facilitating textile recycling processes. Spectral analysis can also utilise the periodicity and orientation inherent in fabric texture, to detect defects. One-dimensional Power Spectral Density analysis using an Auto-Regressive estimation model has been proposed to differentiate normal from defective textures in woven fabrics \cite{bu2010detection}.

The integration of AI with spectral analysis represents a significant advancement, offering enhanced capabilities for precise identification and classification of textiles, thus addressing the evolving challenges in industrial applications. AI can be applied to perform feature extraction, compression, and dimensional reduction of the data in raw spectra \cite{yang2019deep}, keeping the most relevant information, reducing computational complexity, and enhancing the efficiency of subsequent processing. Convolutional neural networks (CNNs) have emerged as a powerful tool in medical applications, enabling tasks such as disease diagnosis from medical images \cite{bandy2023intraclass}.  CNNs have also been utilised to improve identification and sorting of waste textile based on near-infrared spectral data \cite{du2022efficient}. By integrating such AI-driven models, the system is capable of real-time classification, while achieving an accuracy above 95\%, in less than 2 seconds per sample. CNNs can also handle composite materials that involve binary mixtures of common textile fibres, processing the related spectral data to identify patterns and extract relevant features for the accurate classification \cite{riba2022post}.

\subsection{Textile Analysis Datasets}

Textile analysis datasets have been introduced, mostly targeting tasks related to defect detection. The AITEX Fabric Image Database \cite{silvestre2019public} is a publicly available dataset comprising 245 images that include 105 defective samples in 12 categories, and 140 defect-free samples. Overall, the database involves 20 different fabric types with a region of interest of 256 x 256 pixels. Focusing on anomaly detection, the MVTec Anomaly Detection dataset \cite{bergmann2019mvtec} consists of 3629 images allocated across 15 distinct categories designated for training and validation, with an additional 1725 images assigned for testing purposes. The training set, includes only defect-free images. In total, five categories of regular and random texture are included. In addition, a comprehensive resource for fabric detection tasks is the ZJU-Leaper dataset \cite{zju-leaper}. This database encompasses 98,777 samples, featuring 27,650 instances with defects and 71,127 normal samples, representing a total of 19 diverse fabric types. 

While fabric defect detection databases are available, the current landscape of publicly available datasets dedicated to textile type recognition appears to be sparse. Several works have developed and annotated custom data in this context, including popular commercial fabrics \cite{da2020innovative} and blended textiles \cite{feng2019cu}. However, the absence of openly available textile material datasets has hindered the development of robust recognition algorithms, capable of generalising in real-world applications of sorting and recycling pipelines. This underscores the need for creation and sharing of such resources to facilitate advancements in this domain of textile analysis.

\begin{figure*}[t!]
    \centering
    \includegraphics[width=0.85\linewidth]{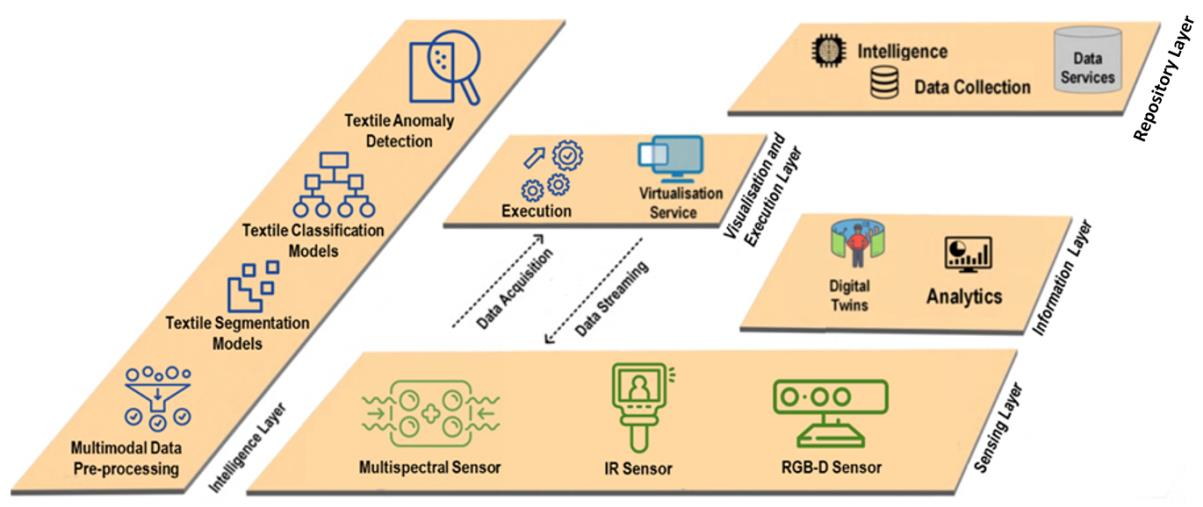}
    \caption{Autonomous textile analysis pipeline architecture.}
    \label{fig:architecture}
\end{figure*}

\subsection{Autonomous Integrated Textile Analysis}

The integration of robotics in cloth manipulation \cite{jimenez2017visual} has recently seen a significant rise, highlighting the potential for enhancing the efficiency and accuracy in textile analysis and sorting by automating handling tasks, while ensuring precision and reducing human errors. Despite these advancements, a comprehensive and holistic approach to automated textile sorting, integrating both robotics and AI with computer vision, is not widely documented in the existing literature. Many studies focus on individual aspects, such as spectral analysis or AI-driven classification, rather than a cohesive integration of these technologies into an automated textile sorting pipeline. In the near future, the potential of 6G on the Internet of Things \cite{spyridis2021towards} could empower these pipelines with even greater capabilities, by embedding sensor technology that could allow real-time data collection on various textile properties. This data, processed through ultra-fast 6G networks, could further refine AI algorithms for automatic material identification and optimise robotic manipulation based on specific fabric characteristics.

\section{Methodology}
\label{sec:methodology}

In this section we outline the summary of our developed 
approach. The methodology is divided into the system architecture, the classification method, including the dataset and training, and the Digital Twin system.

\subsection{Architecture}
The architecture of the autonomous textile analysis pipeline consists of five building blocks:

\begin{enumerate}
    \item Intelligence Layer
    \item Sensing Layer
    \item Information Layer
    \item Visualisation and Execution Layer
    \item Repository Layer
\end{enumerate}

The Intelligence Layer functions as a central hub for the aggregation of data and information derived from the Information Layer, incorporating additional contextual insights such as metadata sourced from the underlying infrastructure and the Repository Layer. Furthermore, this layer establishes a seamless connection with the Sensing Layer to acquire multispectral or other relevant multimodal data, using a spectral camera and other sensors. The collected data is then utilised to execute AI-driven tasks, employing computer vision and machine learning methodologies for the identification, classification, and segmentation of the textile materials.

The outcomes, along with relevant model information, are subsequently transmitted to the Repository Layer, facilitating the storage of obtained results for subsequent analysis. The Information Layer encompasses a comprehensive suite of predictive analytics underpinned by the Digital Twin system and meta-information. Ultimately, the Visualisation and Execution Layer undertakes the presentation of outcomes and all related information, by integrating system management processes and providing visual representations. The overall architecture is demonstrated in Figure \ref{fig:architecture}.

\subsection{AI Classification}

\subsubsection{Training dataset}

The employed dataset comprises cloth images, sourced from \cite{clark2023fabric}. The images were annotated using a sensor capable of discriminating among nine distinct categories of apparel. The data acquisition process employed a HinaLea camera, which captured imagery across 151 spectral bands. These bands span the wavelength range of 950 nm to 1700 nm, with increments of 5nm, encompassing both near-infrared and infrared spectral regions. Data across five material types, including cotton, polyester, silk, wool, and viscose were provided, as presented in Figure \ref{fig:dataset}.

\begin{figure}[b]
    \centering
    \includegraphics[width=\linewidth]{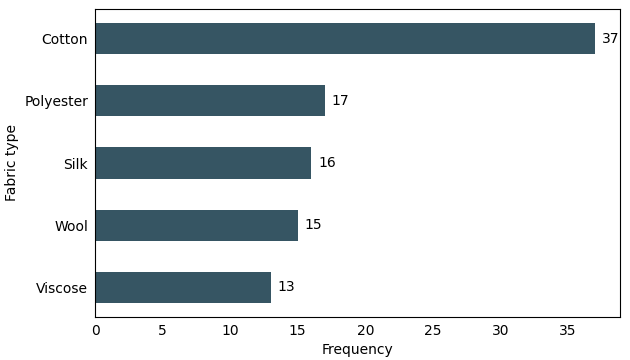}
    \caption{Dataset class distribution.}
    \label{fig:dataset}
\end{figure}



\subsubsection{Convolutional neural networks}

At the core of our textile recognition pipeline CNNs play a central role, utilising their intrinsic capability to proficiently capture spatial hierarchies and patterns in data. Textile fabrics often exhibit complex visual structures with repeating patterns and textures. CNNs excel at learning hierarchical features through the use of convolutional layers, which apply filters to small local regions of the input data, allowing them to detect low-level features like edges and textures. Their ability to recognise patterns irrespective of their spatial positioning make them appropriate for real-world use cases, in which shifts and rotations in images are prevalent.

\begin{table}[t]
    \caption{CNN model summaries.}
    \renewcommand{\arraystretch}{2.0}
    \begin{tabularx}{\linewidth}{|X|X|X|}
      \hline
      \textbf{Model}  & \textbf{Layers \#}  & \textbf{Parameters \#} \\
      \hline
      EfficientNet B6 & \textasciitilde 600 & \textasciitilde 41 million \\
      \hline
      ResNest-101     & \textasciitilde 600 & \textasciitilde 46 million \\
      \hline
      Medium Custom   & 22    & \textasciitilde 1.5 million \\
      \hline
      Simple Custom   & 10    & \textasciitilde 1.5 million \\
      \hline
    \end{tabularx}
    \label{tab:models}
\end{table}

In our study we investigated two custom architectures, and two state-of-the-art pretrained models, fine-tuned in our specific dataset. For the pretrained models, EfficientNet \cite{tan2019efficientnet} and ResNest \cite{zhang2022resnest} were selected, due to their wide success across a variety of image classification tasks. A summary of each model is presented in Table \ref{tab:models}.

\subsection{Digital Twin System}

\begin{figure}[b!]
    \centering
    \includegraphics[width=\linewidth]{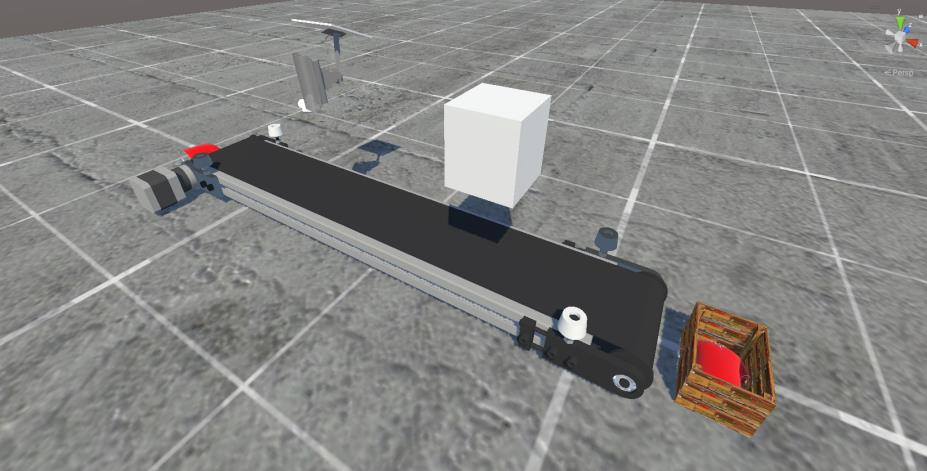}
    \caption{Digital Twin visualisation.}
    \label{fig:dt}
\end{figure}

The integration of the Digital Twin is a critical aspect of our autonomous textile sorting pipeline, as is serves as the virtual counterpart to the physical sorting system, combining real-time data, predictive models, and simulations to replicate and enhance the system dynamics. The DT enables predictive analysis for proactive issue identification, algorithm optimisation, and parameter fine-tuning, thus facilitating superior efficiency and accuracy in the associated textile classification. 

Moreover, the DT functions as a controlled training ground for the CNN models, allowing iterative testing and refinement without constant adjustments to the physical system. Its role extends to fault detection, providing a versatile platform for extensive scenario testing, and ultimately elevating the robustness and reliability of the autonomous textile sorting processes. Table \ref{tab:dt} outlines the main parameters of the DT. For each parameter, there is a setting with respect to the simulated percentage error, ranging from 0-100 \%. A visualisation of the DT is presented in Figure \ref{fig:dt}.


\begin{table}[t]
    \caption{Digital twin key parameters.}
    \renewcommand{\arraystretch}{2.0}
    \begin{tabularx}{\linewidth}{|X|X|}
      \hline
      \textbf{Parameter}        & \textbf{Range}  \\
      \hline
      Conveyor belt speed       & 1-5  \\
      \hline
      Robotic arm speed         & 1-5  \\
      \hline
      Camera capture time       & 3-8  \\
      \hline
      Laser segmentation speed  & 1-5  \\
      \hline
    \end{tabularx}
    \label{tab:dt}
\end{table}

\section{Evaluation}
\label{sec:evaluation}

This section examines the performance and efficacy of the introduced autonomous textile sorting pipeline. It presents the key metrics used to provide insights into the model's ability to accurately categorise textiles and their respective results. IN addition, the evaluation includes the integration of the Digital Twin and investigates its role in enhancing the system's efficiency and reliability.

\subsection{Classification Results}

In assessing the classification performance, precision, F1 score, and accuracy serve as the key benchmarks.

\begin{equation}
    \text{Precision} = \frac{TP}{TP + FP},
\end{equation}
 

\begin{equation}
    \text{Accuracy} = \frac{TP + TN}{TP + TN + FP + FN},
\end{equation}

\noindent where $TP$, $TN$, $FP$, and $FN$ are the true positive, true negative, false positive, and false negative samples respectively.

\begin{equation}
    \text{F1 Score} = \frac{2 \cdot \text{Precision} \cdot \text{Recall}}{\text{Precision} + \text{Recall}}
\end{equation}

\begin{table}[b]
    \caption{Textile classification results.}
    \renewcommand{\arraystretch}{1.3}
    \begin{tabularx}{\linewidth}{C{2.7cm}C{1.5cm}C{1.5cm}C{1.5cm}}
    \toprule
    \textbf{Model}  & \textbf{Accuracy}  & \textbf{Precision} & \textbf{F1 Score} \\
    \toprule
    EfficientNet B6 &  0.242             &  0.219             & 0.195             \\
    ResNest-101     &  0.586             &  0.670             & 0.618             \\
    Medium Custom   &  0.393             &  0.078             & 0.113             \\
    Simple Custom   &  0.363             &  0.082             & 0.114             \\
    \bottomrule
    \end{tabularx}
    \label{tab:classification_results}
\end{table}


PUT NEW ONE

\begin{figure}[t!]
    \centering
    \includegraphics[width=0.95\linewidth]{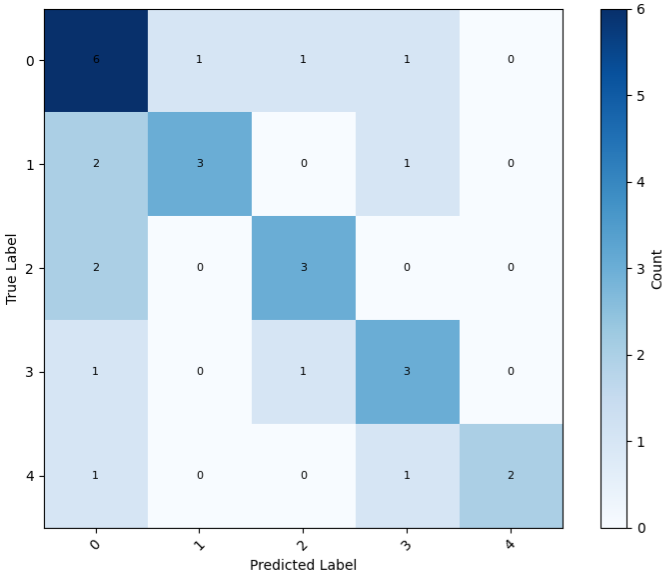}
    \caption{ResNest confusion matrix.}
    \label{fig:c2}
\end{figure}

In addition to the above metrics, ROC-AUC and confusion matrices are also considered. ROC-AUC describes the area under a Receiver Operating Characteristic curve. It provides a single value that represents the ability of the model to distinguish between positive and negative instances. The confusion matrix describes the $TP$, $TN$, $FP$, and $FN$ through a visual representation.

The results of the models across the key metrics are outlined in Table \ref{tab:classification_results}. The scores highlight suboptimal performance across all models, with the exception of the RseNest which was fine-tuned after data augmention. The EfficientNet B6 model demonstrates limited accuracy at 0.242, while precision and F1 scores also remain modest. The Medium Custom and Simple Custom models, although displaying higher accuracy, exhibit notably low precision and F1 scores. The ResNest model was able to achieve the highest accuracy, reaching 0.586, with more robust precision and F1 scores of 0.670 and 0.618 respectively. Overall, these results underscore the need for further refinement and optimisation of the classification models to enhance their effectiveness in accurately categorising textile fabrics within the sorting system. The confusion matrix corresponding to the ResNest model is depicted in Figure \ref{fig:c2}. In the matrix, labels 0-4 correspond to cotton, polyester, wool, silk, and viscose. Figures \ref{fig:roc} and \ref{fig:roc_custom} show the ROC-AUC curves for all the models. The values are mostly in the range of 0.5, again leaving room for improvement.

\begin{figure}[t]
    \centering
    \subfigure[EfficientNet]{
      \includegraphics[width=0.46\linewidth]{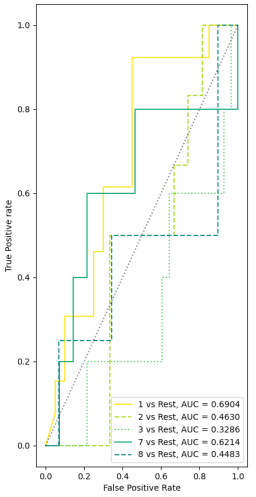}
    }
    \hfill
    \subfigure[ResNest]{
      \includegraphics[width=0.46\linewidth]{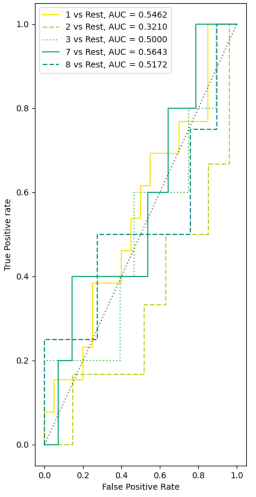}
    }
    \caption{ROC-AUC curves for the EfficientNet and ResNest models.}
    \label{fig:roc}
\end{figure}

\begin{figure}[t]
    \centering
    \subfigure[Medium custom]{
      \includegraphics[width=0.46\linewidth]{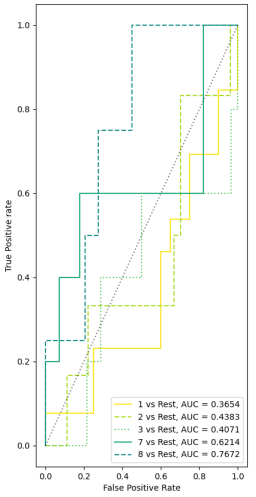}
    }
    \hfill
    \subfigure[Simple custom]{
      \includegraphics[width=0.46\linewidth]{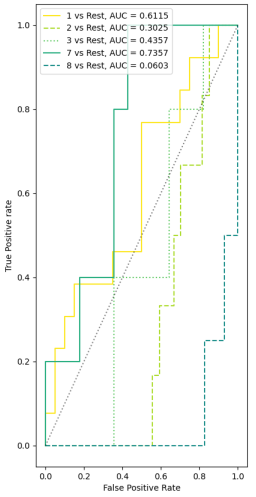}
    }
    \caption{ROC-AUC curves for the Medium and Simple custom models.}
    \label{fig:roc_custom}
\end{figure}

\subsection{Digital Twin Results}

For the evaluation of the Digital Twin system we configured the conveyor belt and robotic arm speed to 5, camera capture speed to 3, and laser segmentation speed to 5. Three experiments were conducted with 10, 12, and 14 clothes assigned for analysis by the system. Each instance was executed 10 times with an 8\% error chance. The average outcomes of the experiments are presented in Table \ref{tab:dt_results}. Note, that the camera capture time metric includes the AI inference time. These results underscore the DT's efficacy in simulating and evaluating the dynamics of our textile sorting system, providing valuable insights into its operational efficiency and performance metrics.

The outcomes of this research hold significant promise for broader applications beyond the textile industry. The core technologies employed, such as AI classification and spectral imaging, demonstrate transferability to other sectors that rely on sorting or quality control processes. This includes industries like waste management, where accurate identification of materials is crucial for proper recycling streams, and agriculture, where efficient sorting of crops and produce can optimise yield and quality. Furthermore, the pipeline's architecture is built on Industry 4.0 principles, allowing for scalability. Thus, the system can be readily adapted to handle larger volumes of data and materials, making it suitable for deployment in bigger facilities or for applications with significant throughput. 

In addition, by enabling predictive analysis and continuous algorithm optimisation though the DT system, this approach can significantly improve efficiency and accuracy across various industries.  This translates to potential cost savings, reduced waste, and improved overall operational performance. Additionally, the DT system can minimise the need for physical prototyping, further contributing to sustainability efforts. Finally, the focus on sustainable textile waste management within the pipeline serves as an inspiration for similar green practices in other fields.  By promoting the efficient sorting and potential reuse of materials, this approach aligns with the principles of a circular economy, where resources are kept in use for extended periods.  The success of this pipeline in the textile industry paves the way for the development of analogous sustainable solutions in diverse sectors, contributing to a more environmentally friendly future. 

\begin{table}[t!]
    \caption{Digital Twin results.}
    \renewcommand{\arraystretch}{1.3}
    \begin{tabularx}{\linewidth}{C{3.7cm}|C{1cm}C{1cm}C{1cm}}
    \toprule
    \textbf{Number of clothes}           &  {10}   &  {12}    & {14}    \\
    \hline
    \textbf{Total time}                  &  80.7 s &  80.3 s  & 93.6 s  \\
    \textbf{Conveyor belt time}          &  18.3 s &  12.7 s  & 14.8 s  \\
    \textbf{Robotic arm time}            &  22.4 s &  19.6 s  & 22.8 s  \\
    \textbf{Camera capture time}         &  30 s   &  36 s    & 42 s    \\
    \textbf{Laser segment time}          &  10 s   &  12 s    & 14 s    \\
    \textbf{Green production efficiency} &  75\%   &  75\%    & 75\%    \\
    \bottomrule
    \end{tabularx}
    \label{tab:dt_results}
\end{table}

\section{Conclusion}
\label{sec:conclusion}

This study addresses the critical challenge of escalating textile waste by presenting a cutting-edge autonomous textile analysis pipeline. The integration of robotics, spectral imaging, AI-driven classification, and laser segmentation has been instrumental in enhancing the accuracy, efficiency, and scalability of textile sorting processes. Our architectural framework, characterised by Industry 4.0 principles, facilitates the seamless data exchange and coordination, forming the basis for a sustainable and circular textile waste management paradigm. While early findings highlight the need for refinement of the classification models and the training data, insights provided by the Digital Twin already show the potential of the automated sorting pipeline for a transition to a greener economy. The incorporation of the laser segmentation further optimises the recycling processes by automating the removal of components such as buttons and zippers.

This holistic approach contributes to the development of a more resource-efficient recycling system. The impact of our work extends beyond technological advancements, as it addresses environmental concerns associated with textile waste and supports the fashion industry's shift towards a more positive recycling approach. The proposed pipeline, assisted by the Digital Twin insights and AI features, lays the groundwork for a transformative and sustainable future in textile waste management.

\section*{Acknowledgement}

This work has received funding from Innovate UK under application number 10080773 for the project titled "AI4Fibres – Artificial Intelligence for Textile and Fibres Recycling". The work was also partially supported by the European Union’s Horizon Europe research and innovation programme under grant agreement No. 101097122 (INCODE).

\bibliographystyle{IEEEtran}
\bibliography{ref}

\end{document}